\documentclass{article}

\usepackage{arxiv}

\usepackage[utf8]{inputenc} % allow utf-8 input
\usepackage[T1]{fontenc}    % use 8-bit T1 fonts
\usepackage{hyperref}       % hyperlinks
\usepackage{url}            % simple URL typesetting
\usepackage{booktabs}       % professional-quality tables
\usepackage{amsfonts}       % blackboard math symbols
\usepackage{nicefrac}       % compact symbols for 1/2, etc.
\usepackage{microtype}      % microtypography
\usepackage{lipsum}
\usepackage{graphicx}
\graphicspath{ {./images/} }

\begin{document}

\title{Sarcasm Detection and Quantification in Arabic Tweets}

\author{
 Bashar Talafha \\
  Computer Science Department\\
  Jordan University of Science\\ and Technology\\
  Irbid, Jordan \\
  \texttt{bashartalafha@gmail.com}
   \And
  Muhy Eddin Za'ter \\
  Electrical and Computer\\ Engineering Department\\
  Princess Sumaya University \\for Technology\\
  Amman, Jordan\\
  \texttt{muhizatar95@gmail.com} 
  \And
 Samer Suleiman \\
  Computer Science \\Department\\
  Jordan University of \\ Science and Technology\\
  Irbid, Jordan \\
  \texttt{sameraas@just.edu.jo} 
  \AND
  Mahmoud Al-Ayyoub\\
  Computer Science \\Department\\
  Jordan University of \\Science \\and Technology\\
  Irbid, Jordan \\
  \textit{maalshbool@just.edu.jo}
  \AND
  Mohammed N. Al-Kabi \\
  Computer Science \\Department\\
  Al-Buraimi University\\
  Al Buraimi, Oman \\
  \textit{mohammed@buc.edu.com}
}

\maketitle

\begin{abstract}
The role of predicting sarcasm in the text is known as automatic sarcasm detection. Given the prevalence and challenges of sarcasm in sentiment-bearing text, this is a critical phase in most sentiment analysis tasks. With the increasing popularity and usage of different social media platforms among users around the world, people are using sarcasm more and more in their day-to-day conversations, social media posts and tweets, and it is considered as a way for people to express their sentiment about some certain topics or issues. As a result of the increasing popularity, researchers started to focus their research endeavors on detecting sarcasm from a text in different languages especially the English language. However, the task of sarcasm detection is a challenging task due to the nature of sarcastic texts; which can be relative and significantly differs from one person to another depending on the topic, region, the user's mentality and other factors. In addition to the aforementioned challenges, sarcasm detection in the Arabic language has its own challenges due to the complexity of the Arabic language, such as being morphologically rich, with many dialects that significantly vary between each other, while also being lowly resourced when compared to English. In recent years, only few research attempts started tackling the task of sarcasm detection in Arabic, including creating and collecting corpora, organizing workshops and establishing baseline models. This paper intends to create a new humanly annotated Arabic corpus for sarcasm detection collected from tweets, and implementing a new approach for sarcasm detection and quantification in Arabic tweets. The annotation technique followed in this paper is unique in sarcasm detection and the proposed approach tackles the problem as a regression problem instead of classification; i.e., the model attempts to predict the level of sarcasm instead of binary classification (sarcastic vs. non-sarcastic) for the purpose of tackling the complex and user-dependent nature of the sarcastic text. The humanly annotated dataset will be available to the public for any usage.
\end{abstract}


\section{Introduction}
Sarcasm in defined in the dictionary as the use of irony to make or convey contempt or ridicule \cite{yavanoglu2018sarcasm}. It is a special sort of sentiment that plays a role as an interrupting element, which tends to flip the given text's polarity \cite{wicana2017review}. In other words, sarcasm has a negative intended sentiment, but may not have a negative surface sentiment. A sarcastic sentence may also carry positive sentiment in the surface (for example, ``Visiting dentists is the best!''), or a negative surface sentiment (for example, ``His writing in the novel was in a matter of fact terrible anyway'' as a response to the criticism of a novel author), or no surface sentiment (for example, the idiomatic expression ``and I am the king of the world'' is used to express sarcasm). Nowadays, with the popularity of social media platforms, people are using sarcasm while expressing their opinions on different subjects, topics, products and services including but not limited to political events, sports games and their day-to-day activities and posts on the different social media platforms concerning all type of topics and issues \cite{joshi2017automatic}.

Since sarcasm implies sentiment, detection of sarcasm in a text is crucial to predicting the correct sentiment of the text which in turn makes sarcasm detection an essential tool that has numerous applications in many fields such as security, health, services, product reviews and sales.
Therefore, major research endeavors started to focus on the task of sarcasm detection from text \cite{rajadesingan2015sarcasm, khalifa2019ensemble, sarsam2020sarcasm, wicana2017review, nayel2021machine}. However, sarcasm detection task is an inherently challenging task due to many reasons. First of all, there are not many available labeled resources of data for sarcasm detection, and any available texts that can be collected (for example Tweets) contains many issues such as an evolving dictionary of slang words and abbreviations, and therefore it usually requires many hours of human annotators to prepare the data for any potential usage. Also, the nature of sarcasm detection adds to the challenging part of the task, as sarcasm can be regarded as being relative and differs significantly between people and it depends on many factors such as the topic, region, time, the events surrounding the sentence and the readers/writers mentality and the ; in other words, a sentence that can be found sarcastic by one person, might sound normal to an another person, which will be further discussed and proven during this work.

Since the start of sarcasm detection research, most of the endeavors were concentrated on the English language, with only a few shy attempts on the Arabic language \cite{al2017arabic, abdulla2013arabic}. The work in this paper focuses on Arabic sarcasm detection by creating a new corpus that is humanly annotated and establishes a baseline classifier with a new approach for the both; the annotation of the data and the detection of sarcasm from text.

The paper is presented and organized as follows; Section~\ref{sec:rel} illustrates the related work to Arabic Sarcasm detection task, while Section~\ref{sec:data} describes the dataset collected and annotated in this work. Section~\ref{sec:method} presents the architecture and methodology of the implemented baseline classifier, followed by the experiments section (Section~\ref{sec:exp}). Finally, the conclusions of this work are presented and discussed in Section~\ref{sec:conc}.

\section{Related Work}
\label{sec:rel}
The task of sarcasm detection is relatively a new task in Natural Language Processing (NLP) and Understanding (NLU) when compared to other tasks. It mainly aroused after the huge popularity of social media platforms which made it possible for users to express their points of views on different issues and events sarcastically. Since the detection of sarcasm plays a role in a lot of applications such as sentiment analysis, researchers started to route their efforts into the task of sarcasm detection. However, as it is the case for most NLP and NLU tasks, the Arabic sarcasm detection does not get as much attention and efforts as the English language. Also, alongside the shortage of available resources in Arabic language when compared to the English language, Arabic being morphologically rich and the fact the it contains a lot of significantly different dialects, impose new challenges on Arabic NLP researchers  \cite{farghaly2009arabic, guellil2019arabic}.

There were only a few shy attempts on Arabic sarcasm detection which were mainly focused on creating datasets from tweets and establish a baseline for each created dataset \cite{ghanem2020irony}. In 2017, the work in \cite{karoui2017soukhria} was the first attempt on sarcasm and irony detection in Arabic language in which the authors created a corpus of sarcastic Arabic tweets that are related to politics. The dataset was created by distant supervision as the authors relied on the keywords that are equivalent of sarcasm and irony in Arabic to label the tweets as ironic or sarcastic. The authors in the paper experimented multiple classifiers on the developed dataset such as Support Vector Machines (SVM), Logistic regression, Naiive Bayes and other classifiers.

In \cite{ghanem2019idat}, the authors organized a shared task on Arabic sarcasm detection in which the authors assembled a dataset that consisted of tweets about different topics, where the tweets were labeled by filtering out the tweets with ironic and sarcastic hashtags and then manually labeled. The winning architecture was developed by the authors in \cite{khalifa2019ensemble} in which they implemented an ensemble classifier of XGBoost, random forest and fully connected neural networks while relying on a set of features that consists of sentiment and statistical features, in addition to word n-grams, topic modelling features and word embeddings. Recently, the authors in \cite{farha2020arabic} released a new dataset (ArSarcasm) that consists of 10k tweets while publishing a baseline model based on Bidirectional Long Short Term Memory (BiLSTM) \cite{hochreiter1997long} architecture. Another dataset was prepared with a corpus of ironic Arabic tweets in which the authors filtered out the hashtags related to sarcasm and irony. Another recent study, created a corpus of ironic tweets, namely DAICT \cite{abbes-etal-2020-daict}. To prepare the corpus, the authors followed the same approach used by \cite{ghanem2019idat}. The most recent work on Arabic sarcasm detection namely ArSarcasm-v2 is an extension on the developed in \cite{farha2020arabic}, where the authors also developed a shared task for Arabic sarcasm detection. The dataset consists of the original ArSarcasm, and the data provided in \cite{abbes-etal-2020-daict}, in addition to newly crawled tweets. The final dataset was humanly annotated by native Arabic speakers through crowd sourcing and was released for participants in a workshop. Participating teams in this workshop utilized many of the recent advancements in deep learning such as Bert, Roberta and other state-of-the-art models and paradigms like multi-task learning to enhance the accuracy of sarcasm/irony detection for Arabic language \cite{el2021deep, abuzayed2021sarcasm, elgabry-etal-2021-contextual, husain-uzuner-2021-leveraging, alharbi2021multi}. The following table summarizes the statistics of each developed dataset.

\begin{table}[h]
\centering
\caption{Summary of Developed datasets}
\begin{tabular}{|c|c|c|c|}
\hline
Dataset & Category     & \# of tweets & Sarcastic tweets \\ \hline
1       & Politics     & 5479         & 1733             \\ \hline
2       & US elections & 5030         & 2614             \\ \hline
3       & Tweets       & 10547        & 1682             \\ \hline
4       & Tweets       & 5358         & 4809             \\ \hline
5       & Tweets       & 15,548       & 5491             \\ \hline
\end{tabular}
\end{table}

The work in this paper proposes a new dataset for Arabic sarcasm that was collected from tweets and humanly annotated using a unique methodology. Along with a baseline model that utilizes multi-dialect Arabic Bert and presents a new approach for sarcasm detection by approaching the problem as a regression task instead of classification task as will be further discussed in details in later sections.

\section{Dataset Collection and Annotation}
\label{sec:data}
In this paper, a new corpus is created for Arabic sarcasm detection. The dataset was collected though Twitter API with the language filter set to Arabic. There are four categories of the collected tweets; entertainment, politics, products and services, and finally sports. The dataset has multiple Arabic dialects and in addition to Modern Standard Arabic (MSA). 

Due to the fact that Sarcasm is usually a relative concept and definition, and it significantly differs from one person to another, in addition to the fact that sometimes it depends on the time and events surrounding sentence (tweet). Therefore, each tweet was annotated by 11 different native Arabic speakers and labeled as sarcastic/non-sarcastic by each annotator. This method of labeling provides not only if a sentence is sarcastic or non-sarcastic but also a level of sarcasm depending on the number of annotators labeling a sentence as sarcastic. 

The final dataset consists of 1554 tweets in 1165 of them are labeled as sarcastic.
The following figures illustrate the distribution of data per label and category.

\begin{figure}[h]
\centering
\centerline{\includegraphics[scale=0.45]{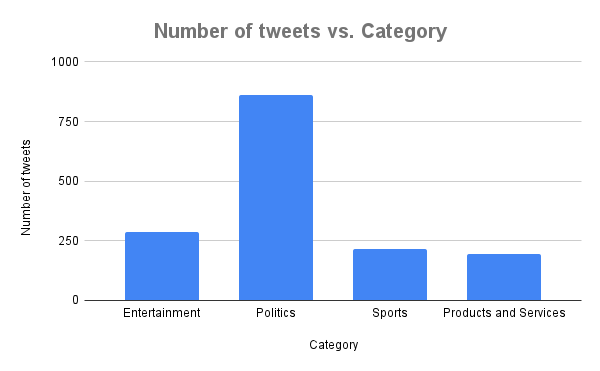}}
\caption{Number of tweets per Category}
\label{fig}
\end{figure}

\begin{figure}[h]
\centering
\centerline{\includegraphics[scale=0.45]{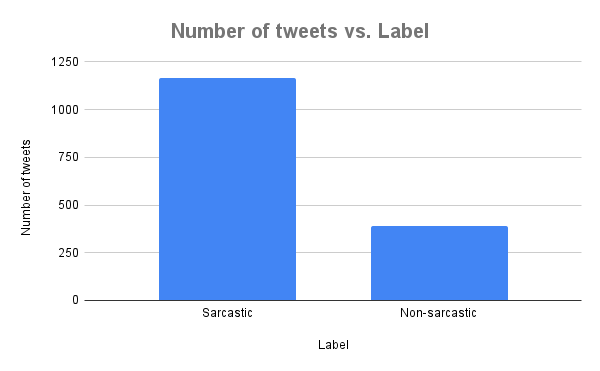}}
\caption{Number of tweets per Label}
\label{fig}
\end{figure}

From the figures above, the majority of the sarcastic tweets crawled were in the category of politics, which can be explained by the fact that people tend to be more sarcastic on political topics.

Table 1 shows a segment of the training data and how it is labeled.

\begin{table*}[h]
\caption{Sample of the datatset}
\centering
\begin{tabular}{|c|c|c|c|c|c|c|c|c|c|c|c|}
\hline
Sentence ID & Ann. 1 & Ann. 2 & Ann. 3 & Ann. 4 & Ann. 5 & Ann. 6 & Ann. 7 & Ann. 8 & Ann. 9 & Ann. 10 & Ann. 11 \\ \hline
1           & Yes    & Yes    & Yes    & No     & Yes    & Yes    & Yes    & Yes    & Yes    & Yes     & Yes     \\ \hline
2           & Yes    & Yes    & No     & No     & No     & No     & No     & No     & No     & No      & Yes     \\ \hline
3           & No     & Yes    & Yes    & No     & No     & Yes    & Yes    & Yes    & No     & Yes     & Yes     \\ \hline
4           & No     & No     & Yes    & Yes    & No     & No     & Yes    & Yes    & No     & No      & No      \\ \hline
5           & No     & No     & No     & No     & No     & No     & No     & No     & No     & No      & No      \\ \hline
\end{tabular}
\end{table*}

From the table above, it can be noticed how sarcasm is a highly relative concept and significantly differs between annotators.

The annotated dataset will be available for public use.

\section{Methodology}
\label{sec:method}
As aforementioned, a new approach is implemented in this work for the task of sarcasm detection from the Arabic language which will be described in the section. 

The implemented approach is divided into 3 stages; the first being data pre-processing and tokenization step, while the second stage consists of extracting vector representations (embeddings) for the training sentences and finally the final stage is the classifier network.

\begin{figure}[h]
\centering
\centerline{\includegraphics[scale=0.45]{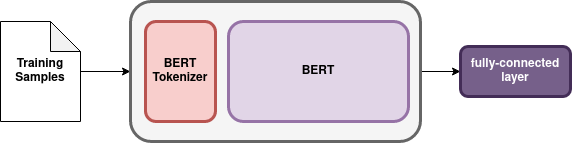}}
\caption{Implemented network Architecture}
\label{fig}
\end{figure}

Looking more closely on the methodology implemented in this paper. For the first stage, the training samples did not undergo any pre-processing and are fed into the BERT tokenizer; specifically ArabicBERT tokenizer \cite{elmadani2020bert}. After the tokenization step, training samples are inserted into a pre-trained multi-dialect Arabic-BERT in  order  to  extract  the  embeddings for each training sample on the sentence level, and hence getting a vector representation for each sentence. These representations are then fed into a simple fully connected neural network, with a sigmoid function as an activation function for the output layer.

\subsection{Multi-Dialect-Arabic-BERT model}
BERT \cite{devlin2018bert} stands for bidirectional encoder representations from Transformers. BERT model’s architecture consists of multiple transformer \cite{vaswani2017attention} encoders for learning contextualized word embedding of a given input text and is trained with a masked language modeling as well as a next sentence prediction objective. BERT based models achieved state-of-the-art results on numerous NLP and NLU tasks.

The BERT model used in the implemented network is Multi-dialect Arabic BERT \cite{talafha2020multi} which was built on the ArabicBERT \cite{farha2020arabic} which is publicly available BERT model trained on around 93 GB of Arabic content crawled from the internet. This model is then fine-tuned on the Nuanced Arabic Dialect Identification (NADI) task \cite{abdul2021nadi} which is the task of identifying 26 Arabic dialect from text, the fine-tuning step was trained with 10 million Arabic sentences crawled from twitter. 

The retrieved vector from the Multi-Dialect-Arabic Bert is then fed into a shallow feed-forward neural classifier implemented for dialect identification. It is worth mentioning that during the fine-tuning process, the loss is propagated back across the entire network, including the BERT encoder.

The fine-tuning step was done on top of ArabicBERT on the 10 million Arabic tweets using an Adam optimizer \cite{kingma2014adam} with a learning rate of $3.75 \times 10^{-5}$
and a batch size of 16 for 3 epochs.

\subsection{Task Classifier}

As aforementioned, the work in this paper approaches the sarcasm detection task as a regression problem instead of a classification problem, in other words, the implemented classifier network predicts the level of sarcasm in a given of a sentence instead of predicting if it is sarcastic or non-sarcastic. A simple classifier is implemented using a feed-forward neural network, the network's hyper-parameters are shown in the table below. The output layer of the network consists of one unit instead of two units, and the activation function is a Sigmoid function instead of Softmax that is used in the conventional classification tasks.

\begin{table}[h]
\centering
\caption{Feed-forward network hyper-parameters}
\begin{tabular}{|l|l|}
\hline
Hyper-parameter     & Value \\ \hline
Size of data        & 1555  \\ \hline
Batch-size          & 8     \\ \hline
Epochs               & 10    \\ \hline
Optimizer           & Adam  \\ \hline
Dropout             & 0.2   \\ \hline
Activation function &  Sigmoid    \\ \hline
Hidden layers       &  2     \\ \hline
Hidden Neurons      &  128    \\ \hline
Output Activation   &  Sigmoid     \\ \hline
Loss                & Mean Squared Error  \\ \hline
\end{tabular}
\end{table}

To summarize the implemented approach, the input sentence is fed into the tokenizer, which in turn is then entered to the trained Multi-Dialect Arabic BERT in order to extract the contextualized embeddings. These embeddings are then fed into a simple feed forward neural network which outputs a number between zero and one that represents the level of sarcasm of the input sentence.

\section{Experiment and Results}
\label{sec:exp}
As mentioned in earlier sections, the sarcasm detection task was treated in this paper as a regression problem instead of a classification problem, and hence the activation function of the output layer  implemented feed-forward neural network ( which consists of one unit only) is Sigmoid instead of the conventional Softmax activation function. Thus the reported results are loss instead of accuracy, F1 score and other classification metrics. The used loss function is the Mean Squared Error (MSE).

As aforementioned, each sentence (tweet) in the dataset was annotated by 11 different Arabic native speaker. The final label represents the level of sarcasm implied in the sentence. The table below shows an example of the training data.

\begin{table}[h]
\centering
\caption{Sample of the labels}
\begin{tabular}{|c|c|}
\hline
Sentence ID & Label \\ \hline
1           & 2/11  \\ \hline
2           & 5/11  \\ \hline
3           & 9/11  \\ \hline
4           & 1     \\ \hline
5           & 0     \\ \hline
6           & 6/11  \\ \hline
\end{tabular}
\end{table}

The Cross-validation method was used to evaluate and divide the data, 10 folds were used; 90\% of the data was used for training where the other 10\% were used for validation. The results of the 10 folds are shown in the table below.

\begin{table}[h]
\caption{Results of Cross-validation}
\centering
\begin{tabular}{|c|c|}
\hline
\textbf{Fold Number} & \textbf{Evaluation loss} \\ \hline
Fold 1               & 0.039962884              \\ \hline
Fold 2               & 0.020321029              \\ \hline
Fold 3               & 0.012043999              \\ \hline
Fold 4               & 0.008851729              \\ \hline
Fold 5               & 0.010085999              \\ \hline
Fold 6               & 0.005922336              \\ \hline
Fold 7               & 0.006725985              \\ \hline
Fold 8               & \textbf{0.003881201}     \\ \hline
Fold 9               & 0.005128059              \\ \hline
Fold 10              & 0.005128059              \\ \hline
\textbf{Final loss}  & \textbf{0.011631458}     \\ \hline
\end{tabular}
\end{table}

\section{Discussion and Conclusion}
\label{sec:conc}
In this paper, the authors attempt to tackle the task of sarcasm detection from Arabic text (tweets) using a different approach than the conventional one. While previous work on all languages tackled the problem as a classification task, in this paper, the task is treated as a regression problem; in other words, the work in this paper tries to quantify the sarcasm in a given tweet instead of detecting it. Due to the complex nature of the sarcasm detection task; which mainly lies in the fact that sarcasm is a relative term between people and is time, events and region dependent. First of all, the tweets crawled in this work are humanly annotated by Arabic native speakers, in which each tweets is annotated by 11 people, and the tweets is labeled as a quantification of sarcasm as mentioned earlier. The dataset will be published and available to the public.
The new method of labeling and training significantly reduces the error of the detection of sarcasm, as instead of predicting if a given text is sarcastic, it predicts the level if sarcasm in this text and hence reduces the error of mis-classification and can in fact provide more information of the sarcasm of the tweet for further applications such as sentiment analysis.
Also, this method of predicting sarcasm tackles the relative nature of sarcasm, and takes into consideration multiple the sarcastic opinion of multiple people. Table V shows the results of the experiments and illustrates that the trained network achieves a good approximation of the humanly annotated level of sarcasm, leveraging the multiple opinions in labeling the data.

The results in this paper will be a benchmark for future experiments on this task and on this dataset (which will be released for public).

\bibliographystyle{unsrt}  
\bibliography{references}

\end{document}